\documentclass{article}

\PassOptionsToPackage{square,comma,numbers}{natbib}
\usepackage[preprint]{nips} %just a preprint

\usepackage{parskip}        % Package to tweak paragraph skipping
\usepackage{tikz}           % Package for drawing
\usepackage{amsmath}
\usepackage{enumitem}
\usepackage{sidecap}
\usepackage{rotating}
\usepackage{float}
\usepackage{arydshln}
\usepackage{multirow}

\usepackage{adjustbox}
\usepackage{changepage}

\usepackage{multirow}
\usepackage{subcaption}

\usepackage[utf8]{inputenc} % allow utf-8 input
\usepackage[T1]{fontenc}    % use 8-bit T1 fonts
\usepackage{url}            % simple URL typesetting
\usepackage{booktabs}       % professional-quality tables
\usepackage{amsfonts}       % blackboard math symbols
\usepackage{nicefrac}       % compact symbols for 1/2, etc.
\usepackage{microtype}      % microtypography

\definecolor{navyblue}{rgb}{0.0, 0.0, 0.5}
\definecolor{yelloworange}{rgb}{1.0, 0.43, 0.0}

\usepackage[
    colorlinks = true,
    linkcolor = black,
    urlcolor  = navyblue,
    citecolor = black,
    anchorcolor = blue
]{hyperref}

\graphicspath{{figures/}}

\title{Learning structures of the French clinical language: development and validation of word embedding models using 21 million clinical reports from electronic health records}

\author{
    Basile Dura \\
    Innovation and Data, IT Department \\
    Assistance Publique – Hôpitaux de Paris\\
    \texttt{basile.dura-ext@aphp.fr}
    \And
    Charline Jean \\
    Innovation and Data, IT Department \\
    Assistance Publique – Hôpitaux de Paris\\
    \texttt{charline.jean-ext@aphp.fr}
    \AND
    Xavier Tannier, PhD\\
    Sorbonne Université, Inserm, Université Sorbonne Paris Nord\\
    Laboratoire d’Informatique Médicale et d’Ingénierie des Connaissances pour la e-Santé\\
    \texttt{xavier.tannier@sorbonne-universite.fr}
    \AND
    Alice Calliger \\
    Innovation and Data, IT Department \\
    Assistance Publique – Hôpitaux de Paris\\
    \texttt{alice.calliger-ext@aphp.fr}
    \And
    Romain Bey, PhD \\
    Innovation and Data, IT Department \\
    Assistance Publique – Hôpitaux de Paris\\
    \texttt{romain.bey@aphp.fr}
    \And
    Antoine Neuraz, MD, PhD \\
    Sorbonne Université, Inserm, Centre de Recherche des Cordeliers\\
    Biomedical Informatics Department, Hôpital Necker-Enfants Malades\\
    Assistance Publique – Hôpitaux de Paris\\
    Team HeKA, INRIA\\
    \texttt{antoine.neuraz@aphp.fr}
    \AND
    Rémi Flicoteaux, MD, PhD \\
    Medical Information Department\\
    Assistance Publique – Hôpitaux de Paris\\
    \texttt{remi.flicoteaux@aphp.fr}
}

\date{\today}

\begin{document}
\maketitle

% \newpage

\begin{abstract}

\textbf{Background}

Clinical studies using real-world data may benefit from exploiting clinical reports, a particularly rich albeit unstructured medium. To that end, natural language processing can extract relevant information. Methods based on transfer learning using pre-trained language models have achieved state-of-the-art results in most NLP applications; however, publicly available models lack exposure to speciality-languages, especially in the medical field.

\textbf{Objective}

We aimed to evaluate the impact of adapting a language model to French clinical reports on downstream medical NLP tasks.

\textbf{Methods}

We leveraged a corpus of 21M clinical reports collected from August 2017 to July 2021 at the Greater Paris University Hospitals (APHP) to produce two CamemBERT architectures on speciality language: one retrained from scratch and the other using CamemBERT as its initialisation. We used two French annotated medical datasets to compare our language models to the original CamemBERT network, evaluating the statistical significance of improvement with the Wilcoxon test.

\textbf{Results}

Our models pretrained on clinical reports increased the average F1-score on APMed (an APHP-specific task) by 3 percentage points to 91\%, a statistically significant improvement. They also achieved performance comparable to the original CamemBERT on QUAERO. These results hold true for the fine-tuned and from-scratch versions alike, starting from very few pre-training samples.

\textbf{Conclusions}

We confirm previous literature showing that adapting generalist pre-train language models such as CamenBERT on speciality corpora improves their performance for downstream clinical NLP tasks. Our results suggest that retraining from scratch does not induce a statistically significant performance gain compared to fine-tuning.

\end{abstract}


\textbf{Abbreviations}

\begin{table}[!ht]
    \centering
    \begin{tabular}{ll}
        \toprule
        \textbf{Abbreviation} & \textbf{Meaning} \\
        \midrule
        \multirow{2}{*}{APHP} & Assistance Publique – Hôpitaux de Paris \\ 
        & (Greater Paris University Hospitals)  \\
        RWD & Real-World Data \\
        EHR & Electronic Health Record \\
        ML & Machine Learning \\
        NLP & Natural Language Processing \\
        BERT & Bidirectional Encoder Representations from Transformers \\
        CDW & Clinical Data Warehouse \\
        EDS & Entrepôt des Données de Santé, APHP's CDW \\
        NER & Named Entity Recognition \\
        \bottomrule \\
    \end{tabular}
\end{table}

\newpage

\section{Introduction}
\label{sec:intro}

Medical studies using real-world data (RWD) may benefit from exploiting clinical reports, a rich albeit unstructured part of electronic health records (EHR) collected during care episodes. These reports may contain relevant information that is scarce in structured EHR: by some estimates, up to 80\% of entities found in clinical reports are absent from other media \cite{raghavan_how_2014}.

In this context and given the scale of data to analyse, natural language processing (NLP) methods are needed to extract meaningful medical information from this unstructured medium, and help address challenges such as automatic detection of adverse drug reaction, clinical trial eligibility or identification of temporal associations \cite{kreimeyer_natural_2017}.

Initially bound to purely rule-based methods, the NLP field has been shifting towards machine learning (ML) algorithms that can detect patterns automatically. The most recent techniques rely on a first processing stage to represent free-text data into machine-readable input using models known as word embeddings, whose goal is to provide a vector representation that conveys as much semantic and syntactic information as possible.

Methods such as GloVe \cite{pennington_glove_2014}, Word2Vec \cite{mikolov_efficient_2013} or fastText \cite{bojanowski_enriching_2017} can learn meaningful static representations for words, but novel embeddings algorithms like ELMo \cite{peters_deep_2018} and XLNet \cite{yang_xlnet_2020} have since been proposed to include contextual information in the embeddings. Introduced by Delvin et al, the Bidirectional Encoder Representations from Transformers (BERT) \cite{devlin_bert_2019} proposes an efficient method outputting rich representations for words based on their context that consistently demonstrates state-of-the-art performance in most NLP applications. In French, FlauBERT \cite{le_flaubert_2020} and CamemBERT \cite{martin_camembert_2020} are trained on general-purpose French-language documents crawled from the Internet.

Using transfer-learning, such pre-trained models can serve as a basis for a variety of NLP tasks. In the context of a clinical data warehouse (CDW), an ecosystem of researchers and clinicians may rely on a shared pre-trained language model, and fine-tune it on their specific tasks.

Previous work has shown that using specialty language for training BERT-based models can widely increase performances \cite{lee_biobert_2019,alsentzer_publicly_2019}: specialty languages and clinical reports in particular follow a distinct syntax and vocabulary, such that training a model to learn these specificities can represent an advantage. Moreover, Martin et al \cite{martin_camembert_2020} have determined that a model trained on a carefully selected subcorpus could achieve comparable results despite using less than 10\% of the original training data.

In this work, we leverage the CDW of the Greater Paris University Hospitals (Entrepôt des Données de Santé, EDS) to confirm whether there is significant advantage to using a word embedding model specifically trained on French clinical reports for clinical NLP tasks, and address the following questions:

\begin{enumerate}
    \item Is there an advantage to retraining from scratch, as opposed to fine-tuning an existing model, given the excess computational toll and environmental footprint?
    \item How many training steps and examples are necessary to learn useful knowledge about the speciality language?
\end{enumerate}

\section{Methods}
\label{sec:methods}

This study followed the RECORD reporting guideline \cite{benchimol_reporting_2015}; the checklist is available in the appendix \ref{app:reproducibility/RECORD}.

\subsection{Dataset}
\label{sec:methods/datasets}

The EDS contains data collected in the EHR of 39 hospitals from the greater Paris area and relative to 11M patients, including 80M clinical text reports.

The training corpus for this work consists of clinical reports gathered between August 2017 and July 2021. Documents are pseudonymised \cite{paris_desidentification_2019} to preserve privacy, by replacing directly identifying entities with fake entities.

Reports containing less than 20 characters were removed and the corpus was resampled to limit the influence of over-represented report types (e.g. prescriptions, consultation or imaging reports).

We pre-processed selected documents with \texttt{EDS-NLP} \cite{dura_eds-nlp_2022} and \texttt{spaCy} \cite{honnibal_spacy_2020} by removing textual pollution, such as administrative information shared by a large proportion of the reports, which could skew the distribution seen by the model (see appendix \ref{app:dataset} for details). Although clinical reports may contain other forms of duplicate information \cite{digan_evaluating_nodate}, we remained conservative and did not push the pre-processing further.

This study was authorised by the EDS institutional review board (IRB 00011591, project CSE-19-20). The EDS is approved by the French national data protection agency (CNIL, decision 1980120).

\subsection{Models}
\label{sec:methods/models}

We used the architecture of CamemBERT-base for all the experiments and compared two training strategies: fine-tuning or retraining it “from scratch”. In what follows, we focus on two models which we compare to the freely-accessible CamemBERT-base model:

\begin{enumerate}
    \item \textit{EDS-fine-tuned}, a version fine-tuned on our clinical documents but using the original weights as the initialisation.
    \item \textit{EDS-from-scratch}, a version trained from the ground up. This approach lets us retrain a domain-specific tokenizer.
\end{enumerate}

Since most reports go over the 512-token limit imposed by the BERT architecture, we decided to segment documents into 128-token-long sequences.

\subsection{Training}
\label{sec:methods/training}

The dataset was split into training (19.6M documents) and validation subsets (1M documents). We pre-trained seven models:

\textbf{EDS-from-scratch} was initialised with random weights. We followed CamemBERT’s training procedure, and ran the experiment for twelve full epochs, totalling more training steps to compensate for the smaller batch size.

\textbf{EDS-fine-tuned} used CamemBERT-base as its initialisation point, and was trained for one epoch on the full dataset. We also trained five other versions to estimate the impact of the number of training samples, using 100K, 300K, 1M, 3M, 10M and 21M documents. We sampled the documents uniformly from the training dataset described earlier. Every model used in this comparison was trained with the same number of steps, corresponding to one full epoch on 21M documents.

We relied on the transformers library by \texttt{HuggingFace} \cite{wolf_transformers_2020}, \texttt{Pytorch} \cite{noauthor_pytorchpytorch_2022} and \texttt{Pytorch-Lightning} \cite{falcon_pytorch_2019} for our code base.

\subsection{Validation}
\label{sec:methods/validation}

\subsubsection{Intrinsic validation}
\label{sec:methods/validation/intrinsic}

We validated our models using their perplexity measured on a held out validation set, and investigated the influence of the tokenization step. We compared the distribution of tokenized sequence lengths to evaluate whether the tokenizer had learnt some useful information about the clinical vocabulary.

\subsubsection{Extrinsic validation}
\label{sec:methods/validation/extrinsic}

We validated our models on two named entity recognition (NER) tasks, see appendix \ref{app:tasks} for detail:

\begin{itemize}
    \item APMed \cite{neuraz_natural_2020,jouffroy_hybrid_2021}: a corpus for extracting drug related information in clinical reports in French.
    \item QUAERO \cite{neveol_quaero_nodate}, a compilation of two French corpora annotated to ten types of clinical entities:
    \begin{itemize}
        \item EMEA includes long texts containing information on marketed drugs from the European Medicines Agency;
        \item MEDLINE regroups titles of research articles.
    \end{itemize}
\end{itemize}

Every task was framed as a token classification problem, using IOB2 notation. We used the same architecture for every test, and trained the models in depth during fine-tuning on the downstream task. We added a classification head consisting of:

\begin{itemize}
    \item A fully-connected hidden layer with ReLU activation;
    \item A fully-connected output layer.
\end{itemize}

Experiments were reproduced ten times with different random initialisations, to obtain a confidence interval around the results. We used seqeval \cite{nakayama_seqeval_2018} to compute the micro-averaged F1-score, and we evaluated the statistical significance using a Wilcoxon signed-rank test. All tests were 2-sided and p-values were considered statistically significant when lower than 0.05.

\section{Results}
\label{sec:results}

\subsection{Training}
\label{sec:results/training}

Training \textit{EDS-from-scratch} on 21M reports for 12 epochs took 25 days on 8 Tesla V100 GPUs. Each version of EDS-CamemBERT-fine-tuned was trained for 2 days on the same setup. Total carbon emissions were estimated using the MachineLearning Impact calculator \cite{lacoste_quantifying_2019} at respectively 10 and 110 kgCO2eq for each version of \textit{EDS-fine-tuned} and \textit{EDS-from-scratch}.

\subsection{Intrinsic validation}
\label{sec:results/validation-intrinsic}

The median number of tokens needed to represent one document was 1724 for CamemBERT’s original tokenizer, and 1500 using our EDS-specific tokenizer.

The models’ loss on unseen data was still decreasing at the end of training (see appendix \ref{app:training}).

\subsection{Extrinsic validation}
\label{sec:results/validation-extrinsic}

\subsubsection{Comparison with CamemBERT-base}
\label{sec:results/validation-extrinsic/comparison}

We compared the transfer-learning capabilities of our models with CamemBERT-base, and recapitulated the results in Table \ref{tab:perf-extrinsic}.

\begin{table}[h]
    \caption{Performance of our models on multiple extrinsic tasks compared to CambemBERT-base. Results are formatted as mean (+/-std).\\ *: significantly different from from CamemBERT-base (p-value: $p<0.05$)}
    \label{tab:perf-extrinsic}
    \begin{adjustwidth}{-.5in}{-.5in}  
        \centering
        \begin{tabular}{ ccccc }
            \toprule
            \multirow{2}{*}{\textbf{Model}} & \multirow{2}{*}{\textbf{APMed (F1-score)}} & \multicolumn{3}{ c }{\textbf{QUAERO (F1-score)}} \\
            & & \textbf{EMEA} & \textbf{MEDLINE} & \textbf{Total} \\
            \midrule
            \textit{EDS-fine-tuned} &
            .902 (±0.003)* &
            .729 (±0.008) * &
            .597 (±0.007) * &
            \textbf{.655 (±0.007)} \\ 
            \textit{EDS-from-scratch} &
            \textbf{.908 (±0.005)} * &
            .693 (±0.012) * &
            \textbf{.601 (±0.01)} * &
            .642 (±0.007) * \\ 
            CamemBERT-base &
            .866 (±0.007) &
            \textbf{.737 (±0.006)} &
            .584 (±0.004) &
            .651 (±0.004) \\
            
            Best QUAERO \cite{neveol_clinical_nodate} model &
            &
            .749 &
            .698 & \\
            \bottomrule \\
            
        \end{tabular}
    
    \end{adjustwidth}
\end{table}

The results on APMed, the EDS-specific dataset, show a statistically significant improvement when using re-trained language models (maximum $p = 4 \cdot 10^{-3}$). However, the difference between \textit{EDS-fine-tuned} and \textit{EDS-from-scratch} is not significant ($p = 0.1$).

On QUAERO, \textit{EDS-fine-tuned} performed better than \textit{EDS-from-scratch} overall ($p = 4 \cdot 10^{-3}$).

\subsubsection{Impact of the number of training steps and training examples}
\label{sec:results/validation-extrinsic/impact}

\begin{figure}[ht]
    \centering
    \includegraphics[width=.7\textwidth]{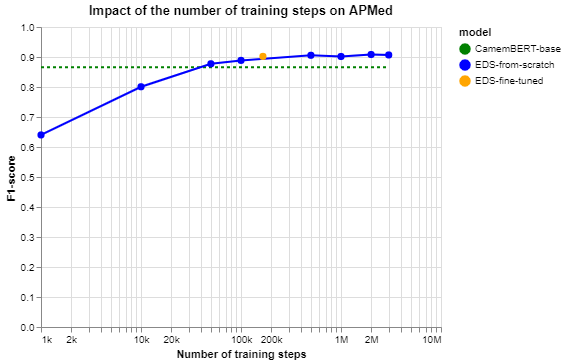}
    \caption{Impact of the number of training steps for \textit{EDS-from-scratch} on the F1-score for the APMed task (logarithmic scale on the horizontal axis). CamemBERT-base is presented as reference (not re-trained). Each model goes through 5.6M documents and emits roughly 4 kgCO2eq every 100k steps.}
    \label{fig:training-steps}
\end{figure}

Figure \ref{fig:training-steps} investigates the impact of the number of training steps performed on \textit{EDS-from-scratch} on its performance on the APMed dataset.

Moreover, Table \ref{tab:num-train-examples} shows the impact of the number of examples when fine-tuning CamemBERT-base.

\begin{table}[!ht]
    \centering
    \caption{Performance on APMed with respect to the number of the training examples}
    \label{tab:num-train-examples}
    \begin{tabular}{cc}
        \toprule
        \textbf{Training examples} & \textbf{APMed (F1-score)} \\
        \midrule
        100K & .900 (±0.004) \\
        300K & .900 (±0.006) \\
        1M & .904 (±0.004) \\
        3M & .904 (±0.003) \\
        10M & .901 (±0.004) \\
        21M (\textit{EDS-fine-tuned}) & .902 (±0.003) \\
        \bottomrule \\
    \end{tabular}
\end{table}

\section{Discussion}
\label{sec:discussion}

In this work, we investigated the impact of pre-training a BERT-based language model on clinical reports by comparing the performance on two medical down-stream NER tasks.

Our results on the APMed corpus confirm previous literature findings that pre-training on speciality language leads to a statistically significant performance improvement.

We also evaluated our models outside the EDS context to check for non-regression, namely on the QUAERO corpora, and observed that \textit{EDS-fine-tuned} fared comparably to CamemBERT-base on this non-clinical dataset. However, BERT-based methods achieved much lower performance than QUAERO’s rule-based laureates on the MEDLINE subcorpus. We posit that the fine-tuned model retains sufficient general language knowledge to keep relatively high performances on a non-clinical task, and that the short length of MEDLINE’s examples hinders contextual methods, although further validation is needed.

What is more, even on the EDS-specific APMed task, we found no significant advantage to using a model pre-trained from scratch compared to fine-tuning a generalist model. Although the new tokenizer is better apt to capture the medical vocabulary (the median token sequence length on EDS reports drops by 15\% when using the EDS-specific tokenizer), the overall performance on EDS-specific NER tasks is similar (Table \ref{tab:num-train-examples}). The impact in terms of inference time is to be evaluated in a future work.

Moreover, our study on the impact of the number of training steps and training examples suggests that relatively few samples are required to reach good test performance. Indeed, our model fine-tuned on 100K samples was able to reach performances similar to models trained on the full 21M-report dataset, at a fraction of the computational and environmental toll. This finding opens up a world of possibilities for smaller-scale CDW, which can adapt general-language models to their distribution at a relatively low cost.

\section{Conclusion}
\label{sec:conclusion}

In this work, we propose EDS-CamemBERT, a language modelling neural network adapted to the context of French-speaking clinical data warehouses. We show that fine-tuning state-of-the-art language models on clinical reports improves performances on downstream speciality tasks. We demonstrate that in this setting, training a model from scratch bears little advantages to fine-tuning a general-language model, despite providing a better tuned tokenizer. Finally, we provide evidence that very few samples are needed to achieve a statistically significant gap in performance.

\section*{Acknowledgement}
\label{sec:acknowledgement}

We thank the Greater Paris University Hospitals CDW for its support and the realisation of data management and data curation tasks.

\section*{Conflicts of Interest statement}
\label{sec:conflicts-of-interest}

None declared.

\section*{Data and code sharing}
\label{sec:data-code-sharing}

Access to the Clinical Data Warehouse's raw data can be granted following the process described on its website: eds.aphp.fr. A prior validation of the access by the local IRB is required. In the case of non-APHP researchers, the signature of a collaboration contract is also mandatory.

The source code used for building the dataset and training the models is freely available on APHP’s Github account, distributed under a 3-Clause BSD licence. It is documented, versioned and citable through Zenodo.

\section*{Funding}
\label{sec:funding}

This study has been supported by grants from the APHP Foundation.

% \nocite{*}
\bibliographystyle{unsrtnat}
\bibliography{main.bib}
\newpage

\appendix

\section{Reproducibility}
\label{app:reproducibility}

\subsection{Code availability}
\label{app:reproducibility/code-availability}

The codebase used to train and evaluate our models is saved on Zenodo and accessible on APHP’s Github account, under an open-source (3-Clause BSD) licence.

\subsection{Data sharing}
\label{app:reproducibility/data}

Trained model weights and the training dataset can be accessed after review by APHP’s institutional review board.

\subsection{The RECORD statement checklist}
\label{app:reproducibility/RECORD}

\begin{table}[ht]
    \centering
    \caption{RECORD checklist}
    \label{tab:RECORD}
    \tiny
    \begin{tabular}{
        p{.12\textwidth}
        p{.02\linewidth}
        p{.2\textwidth}
        p{.15\textwidth}
        p{.2\textwidth}
        p{.15\textwidth}
    }
        \toprule
        & \textbf{Item No} 
        & \textbf{STROBE items} 
        & \textbf{Location in manuscript where items are reported} 
        & \textbf{RECORD items} 
        & \textbf{Location in manuscript where items are reported} \\
        \hline
        \textbf{Title and abstract} & & & & & \\ \hline
        & 1 & (a) Indicate the study’s design with a commonly used term in the title or the abstract (b) Provide in the abstract an informative and balanced summary of what was done and what was found & a) Abstract - Methods b) Abstract - Methods and Results & RECORD 1.1: The type of data used should be specified in the title or abstract. When possible, the name of the databases used should be included.  RECORD 1.2: If applicable, the geographic region and timeframe within which the study took place should be reported in the title or abstract.  RECORD 1.3: If linkage between databases was conducted for the study, this should be clearly stated in the title or abstract. & 1.1) Abstract - Methods 1.2) Abstract - Methods 1.3) N/A \\ \hline
        \textbf{Introduction} &  &  &  &  & ~ \\ \hline
        Background rationale & 2 & Explain the scientific background and rationale for the investigation being reported & Section \ref{sec:intro} &  & ~ \\ \hline
        Objectives & 3 & State specific objectives, including any prespecified hypotheses & Section \ref{sec:intro} &  & ~ \\ \hline
        \textbf{Methods} &  &  &  &  & ~ \\ \hline
        Study Design & 4 & Present key elements of study design early in the paper & Sections \ref{sec:methods/datasets}, \ref{sec:methods/models}, \ref{sec:methods/training}, \ref{sec:methods/validation} &  & ~ \\ \hline
        Setting & 5 & Describe the setting, locations, and relevant dates, including periods of recruitment, exposure, follow-up, and data collection & Section \ref{sec:methods/datasets} &  & ~ \\ \hline
        Participants & 6 & (a) Cohort study - Give the eligibility criteria, and the sources and methods of selection of participants. Describe methods of follow-up Case-control study - Give the eligibility criteria, and the sources and methods of case ascertainment and control selection. Give the rationale for the choice of cases and controls Cross-sectional study - Give the eligibility criteria, and the sources and methods of selection of participants  (b) Cohort study - For matched studies, give matching criteria and number of exposed and unexposed Case-control study - For matched studies, give matching criteria and the number of controls per case & a) Section \ref{sec:methods/datasets}  b) NA & RECORD 6.1: The methods of study population selection (such as codes or algorithms used to identify subjects) should be listed in detail. If this is not possible, an explanation should be provided.  RECORD 6.2: Any validation studies of the codes or algorithms used to select the population should be referenced. If validation was conducted for this study and not published elsewhere, detailed methods and results should be provided.  RECORD 6.3: If the study involved linkage of databases, consider use of a flow diagram or other graphical display to demonstrate the data linkage process, including the number of individuals with linked data at each stage. & 6.1) Section \ref{sec:methods/datasets}  6.2) Section \ref{sec:methods/datasets}  6.3) NA \\ \hline
        Variables & 7 & Clearly define all outcomes, exposures, predictors, potential confounders, and effect modifiers. Give diagnostic criteria, if applicable. & NA & RECORD 7.1: A complete list of codes and algorithms used to classify exposures, outcomes, confounders, and effect modifiers should be provided. If these cannot be reported, an explanation should be provided. & NA \\ \hline
        Data sources/ measurement & 8 & For each variable of interest, give sources of data and details of methods of assessment (measurement). Describe comparability of assessment methods if there is more than one group & Section \ref{sec:methods/datasets} &  & ~ \\ \hline
        Bias & 9 & Describe any efforts to address potential sources of bias & Section \ref{sec:methods/datasets} &  & ~ \\ \hline
        Study size & 10 & Explain how the study size was arrived at & Section \ref{sec:methods/datasets} &  & ~ \\ \hline
        Quantitative variables & 11 & Explain how quantitative variables were handled in the analyses. If applicable, describe which groupings were chosen, and why & NA &  & ~ \\ \hline
        Statistical methods & 12 & (a) Describe all statistical methods, including those used to control for confounding (b) Describe any methods used to examine subgroups and interactions (c) Explain how missing data were addressed (d) Cohort study - If applicable, explain how loss to follow-up was addressed Case-control study - If applicable, explain how matching of cases and controls was addressed Cross-sectional study - If applicable, describe analytical methods taking account of sampling strategy (e) Describe any sensitivity analyses & a) Sections \ref{sec:methods/training}, \ref{sec:methods/validation}  b) Section \ref{sec:methods/validation}  c) Section \ref{sec:methods/datasets}  d) NA  e) Section \ref{sec:methods/validation} &  & ~ \\

        \bottomrule
    \end{tabular}
\end{table}

\begin{table}[ht]
    \centering
    \tiny
    \begin{tabular}{
        p{.12\textwidth}
        p{.02\linewidth}
        p{.2\textwidth}
        p{.15\textwidth}
        p{.2\textwidth}
        p{.15\textwidth}
    }
        \toprule
        & \textbf{Item No} 
        & \textbf{STROBE items} 
        & \textbf{Location in manuscript where items are reported} 
        & \textbf{RECORD items} 
        & \textbf{Location in manuscript where items are reported} \\
        \hline

        Data access and cleaning methods &  &  &  & RECORD 12.1: Authors should describe the extent to which the investigators had access to the database population used to create the study population.  RECORD 12.2: Authors should provide information on the data cleaning methods used in the study. & 12.1) Section \ref{sec:methods/datasets} 12.2) Section \ref{sec:methods/datasets} \\ \hline
        Linkage &  &  &  & RECORD 12.3: State whether the study included person-level, institutional-level, or other data linkage across two or more databases. The methods of linkage and methods of linkage quality evaluation should be provided. & 12.3) NA \\ \hline
        \textbf{Results} &  &  &  &  & ~ \\ \hline
        Participants & 13 & (a) Report the numbers of individuals at each stage of the study (e.g., numbers potentially eligible, examined for eligibility, confirmed eligible, included in the study, completing follow-up, and analysed) (b) Give reasons for non-participation at each stage. (c) Consider use of a flow diagram & a) Section \ref{sec:results/training}  b) Appendix \ref{app:dataset/flowchart}  c) Appendix \ref{app:dataset/flowchart} & RECORD 13.1: Describe in detail the selection of the persons included in the study (i.e., study population selection) including filtering based on data quality, data availability and linkage. The selection of included persons can be described in the text and/or by means of the study flow diagram. & 13.1) Section \ref{sec:results/training} and appendix \ref{app:dataset/flowchart} \\ \hline
        Descriptive data & 14 & (a) Give characteristics of study participants (e.g., demographic, clinical, social) and information on exposures and potential confounders (b) Indicate the number of participants with missing data for each variable of interest (c) Cohort study - summarise follow-up time (e.g., average and total amount) & a) Appendices \ref{app:dataset/subsampling}, \ref{app:dataset/pollution}, \ref{app:dataset/description}  b) Appendices \ref{app:dataset/subsampling}, \ref{app:dataset/pollution}  c) NA &  & ~ \\ \hline
        Outcome data & 15 & Cohort study - Report numbers of outcome events or summary measures over time Case-control study - Report numbers in each exposure category, or summary measures of exposure Cross-sectional study - Report numbers of outcome events or summary measures & NA &  & \\ \hline
        Main results & 16 & (a) Give unadjusted estimates and, if applicable, confounder-adjusted estimates and their precision (e.g., 95\% confidence interval). Make clear which confounders were adjusted for and why they were included (b) Report category boundaries when continuous variables were categorized (c) If relevant, consider translating estimates of relative risk into absolute risk for a meaningful time period & a) Section \ref{sec:results/validation-extrinsic}  b) NA  c) NA &  & \\ \hline
        Other analyses & 17 & Report other analyses done—e.g., analyses of subgroups and interactions, and sensitivity analyses & Section \ref{sec:results/validation-extrinsic} &  & ~ \\ \hline
        % \textbf{Discussion} &  &  &  &  & \\ \hline
        % Key results & 18 & Summarise key results with reference to study objectives & Section \ref{sec:discussion} &  & ~ \\ \hline
        % Limitations & 19 & Discuss limitations of the study, taking into account sources of potential bias or imprecision. Discuss both direction and magnitude of any potential bias & Section \ref{sec:discussion} & RECORD 19.1: Discuss the implications of using data that were not created or collected to answer the specific research question(s). Include discussion of misclassification bias, unmeasured confounding, missing data, and changing eligibility over time, as they pertain to the study being reported. & Section \ref{sec:discussion} \\ \hline
        % Interpretation & 20 & Give a cautious overall interpretation of results considering objectives, limitations, multiplicity of analyses, results from similar studies, and other relevant evidence & Section \ref{sec:discussion} &  & ~ \\ \hline
        % Generalisability & 21 & Discuss the generalisability (external validity) of the study results & Section \ref{sec:discussion} &  & ~ \\ \hline
        % Other Information &  &  &  &  & ~ \\ \hline
        % Funding & 22 & Give the source of funding and the role of the funders for the present study and, if applicable, for the original study on which the present article is based & \hyperref[sec:funding]{Funding} &  & ~ \\ \hline
        % Accessibility of protocol, raw data, and programming code &  &  &  & RECORD 22.1: Authors should provide information on how to access any supplemental information such as the study protocol, raw data, or programming code. & \hyperref[sec:data-code-sharing]{Data and code sharing} \\
        
        \bottomrule
    \end{tabular}
\end{table}

\begin{table}[ht]
    \centering
    \tiny
    \begin{tabular}{
        p{.12\textwidth}
        p{.02\linewidth}
        p{.2\textwidth}
        p{.15\textwidth}
        p{.2\textwidth}
        p{.15\textwidth}
    }
        \toprule
        & \textbf{Item No} 
        & \textbf{STROBE items} 
        & \textbf{Location in manuscript where items are reported} 
        & \textbf{RECORD items} 
        & \textbf{Location in manuscript where items are reported} \\
        \hline

        \textbf{Discussion} &  &  &  &  & \\ \hline
        Key results & 18 & Summarise key results with reference to study objectives & Section \ref{sec:discussion} &  & ~ \\ \hline
        Limitations & 19 & Discuss limitations of the study, taking into account sources of potential bias or imprecision. Discuss both direction and magnitude of any potential bias & Section \ref{sec:discussion} & RECORD 19.1: Discuss the implications of using data that were not created or collected to answer the specific research question(s). Include discussion of misclassification bias, unmeasured confounding, missing data, and changing eligibility over time, as they pertain to the study being reported. & Section \ref{sec:discussion} \\ \hline
        Interpretation & 20 & Give a cautious overall interpretation of results considering objectives, limitations, multiplicity of analyses, results from similar studies, and other relevant evidence & Section \ref{sec:discussion} &  & ~ \\ \hline
        Generalisability & 21 & Discuss the generalisability (external validity) of the study results & Section \ref{sec:discussion} &  & ~ \\ \hline
        Other Information &  &  &  &  & ~ \\ \hline
        Funding & 22 & Give the source of funding and the role of the funders for the present study and, if applicable, for the original study on which the present article is based & \hyperref[sec:funding]{Funding} &  & ~ \\ \hline
        Accessibility of protocol, raw data, and programming code &  &  &  & RECORD 22.1: Authors should provide information on how to access any supplemental information such as the study protocol, raw data, or programming code. & \hyperref[sec:data-code-sharing]{Data and code sharing} \\
        
        \bottomrule
    \end{tabular}
\end{table}

\newpage

\section{Training dataset}
\label{app:dataset}

\subsection{Reglementary considerations}
\label{app:dataset/reglementary}

This study has been approved by the institutional review board of the EDS (IRB 00011591, decision CSE 19-20). The EDS, authorised by the French National Data Protection Commission (CNIL, decision 1980120), ensures patients' information through a transparency portal in accordance with the European General Data Protection Regulation. Subjects that objected to the reuse of their data were excluded from this study in accordance with French legislation.

\subsection{Flow chart}
\label{app:dataset/flowchart}

\begin{figure}[ht]
    \centering
    \includegraphics[width=.7\textwidth]{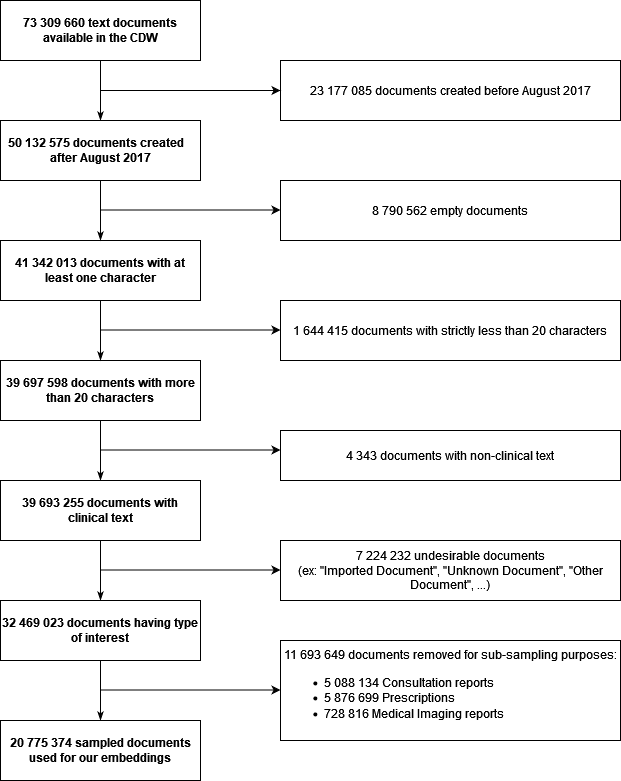}
    \caption{Flow chart detailing the dataset creation}
    \label{fig:dataset-flowchart}
\end{figure}

Figure \ref{fig:dataset-flowchart} presents a flowchart describing the cohord definition.

\subsection{Sub-sampling}
\label{app:dataset/subsampling}

\begin{figure}[ht]
    \centering
    \includegraphics[width=.7\textwidth]{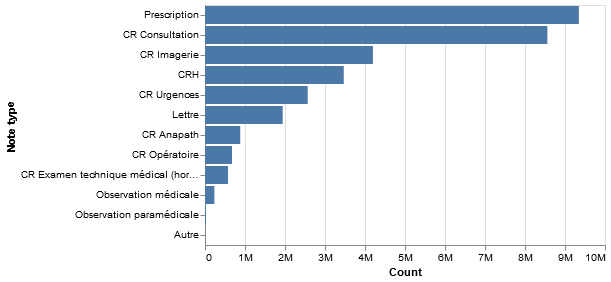}
    \caption{Distribution of meta-types (similar types are regrouped) before hierarchical sampling}
    \label{fig:meta-type-distribution}
\end{figure}

\begin{figure}[ht]
    \centering
    \includegraphics[width=.7\textwidth]{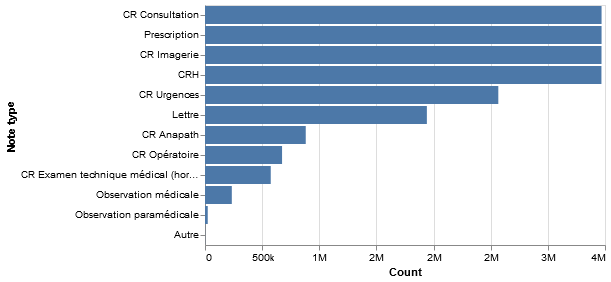}
    \caption{Distribution of meta-types (similar types are regrouped) after hierarchical sampling}
    \label{fig:meta-type-subsampling}
\end{figure}

Figures \ref{fig:meta-type-distribution} and \ref{fig:meta-type-subsampling} show the impact of the sampling method on the distribution of document types. Table \ref{tab:meta-categories-description} proposes a description of the meta-categories retained in this study.

\begin{table}[!ht]
    \centering
    \caption{Description of the meta-categories}
    \label{tab:meta-categories-description}
    \begin{tabular}{ll}
    \toprule
    \textbf{Meta-category} & \textbf{Description} \\
    \midrule
    CR Consultation & Consultation reports \\
    Prescription & Prescriptions \\
    CR Imagerie & Imaging reports \\
    CRH & Hospitalisation reports \\
    CR Urgences & Emergency room reports \\
    Lettre & Lettres \\
    CR Anapath & Anatomopathology reports \\
    CR Opératoire & Operation reports \\
    CR Examen technique médical & Technical examination reports \\
    Observation médicale & Medical observations \\
    Observation paramédicale & Paramedical observations \\
    Autre & Any documents not included in the other categories \\
    \bottomrule \\
    \end{tabular}
\end{table}

\subsection{Removal of administrative pollution}
\label{app:dataset/pollution}

We used \texttt{EDS-NLP} \cite{dura_eds-nlp_2022}, developed at APHP, to tag and remove textual pollution. Such items include the systematic mention of patients’ rights regarding the use of their personal data. \texttt{EDS-NLP} builds on \texttt{spaCy} \cite{honnibal_spacy_2020} to propose open-source pipeline components specifically designed for French-language CDW.

\newpage

\subsection{Description}
\label{app:dataset/description}

Table \ref{tab:cohort-distribution} presents the age and gender distribution of patients within the cohort.

\begin{table}[!ht]
    \centering
    \caption{Distribution of the population within the selected dataset. This is a per-document view, and the age of the patient is calculated relative to the creation date of each individual document.}
    \label{tab:cohort-distribution}
    \begin{tabular}{llll}
    \toprule
    \textbf{Age distribution (\%)} & \textbf{Female} & \textbf{Male} & \textbf{Total} \\
    \midrule
    0-10 & 761 486 (44\%) & 972 365 (56\%) & 1 733 851 (8\%) \\
    10-20 & 592 199 (48\%) & 642 659 (52\%) & 1 234 858 (6\%) \\
    20-30 & 939 914 (56\%) & 751 753 (44\%) & 1 691 667 (8\%) \\
    30-40 & 1 344 079 (59\%) & 934 149 (41\%) & 2 278 228 (11\%) \\
    40-50 & 1 289 314 (54\%) & 1 080 854 (46\%) & 2 370 168 (11\%) \\
    50-60 & 1 523 969 (51\%) & 1 478 293 (49\%) & 3 002 262 (14\%) \\
    60-70 & 1 541 862 (47\%) & 1 768 249 (53\%) & 3 310 111 (16\%) \\
    70-80 & 1 314 876 (46\%) & 1 560 407 (54\%) & 2 875 283 (14\%) \\
    80-90 & 906 750 (53\%) & 807 643 (47\%) & 1 714 393 (8\%) \\
    90+ & 356 648 (67\%) & 177 126 (33\%) & 533 774 (3\%) \\
    Total & 10 571 097 (51\%) & 10 173 498 (49\%) & 20 744 595 \\
    \bottomrule \\
    \end{tabular}
\end{table}

\section{Training}
\label{app:training}

\begin{figure}[ht]
    \centering
    \includegraphics[width=.7\textwidth]{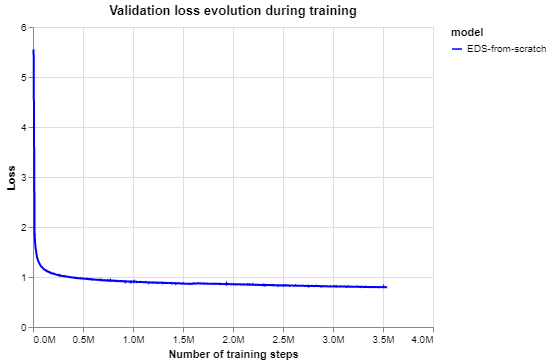}
    \caption{Validation loss evolution during EDS-from-scratch training}
    \label{fig:validation-loss}
\end{figure}

Figure \ref{fig:validation-loss} presents the validation loss during EDS-from-scratch training.

\section{Validation tasks}
\label{app:tasks}

\subsection{APMed}
\label{app:tasks/apmed}

The \textbf{APMed} corpus is a named entity recognition task for the detection and normalisation of drug mentions with their posology. It was developed using clinical reports generated at APHP.

\subsection{QUAERO}
\label{app:tasks/quaero}

The \textbf{QUAERO} dataset, or QUAERO French Medical Corpus, is a compilation of two corpora annotated for medical entities:
\begin{itemize}
    \item EMEA includes long texts containing information on marketed drugs from the European Medicines Agency;
    \item MEDLINE regroups titles of research articles.
\end{itemize}
The documents were annotated for clinical entities defined by a subset of the UMLS Semantic Groups: Anatomy, Chemical and Drugs, Devices, Disorders, Geographic Areas, Living Beings, Objects, Phenomena, Physiology, Procedures.

\end{document}